\documentclass{ieeeaccess}
\usepackage{cite}
\usepackage{amsmath,amssymb,amsfonts}
\usepackage{algorithmic}
\usepackage{graphicx}
\usepackage{textcomp}
\usepackage{subcaption}
\usepackage{multirow}
\def\BibTeX{{\rm B\kern-.05em{\sc i\kern-.025em b}\kern-.08em
    T\kern-.1667em\lower.7ex\hbox{E}\kern-.125emX}}
\begin{document}

\title{Improving colonoscopy lesion classification using semi-supervised deep learning}
\author{\uppercase{Mayank Golhar}\authorrefmark{1},
\uppercase{Taylor L. Bobrow\authorrefmark{2}, \uppercase{MirMilad Pourmousavi Khoshknab}\authorrefmark{3}, \uppercase{Simran Jit}\authorrefmark{3}, \uppercase{Saowanee Ngamruengphong}\authorrefmark{3}, and Nicholas J. Durr}\authorrefmark{1,2}.}
\address[1]{Department of Electrical and Computer Engineering, Johns Hopkins University, Baltimore, 
MD 21218 USA}
\address[2]{Department of Biomedical Engineering, Johns Hopkins University, Baltimore, 
MD 21218 USA}
\address[3]{Division of Gastroenterology, Johns Hopkins Hospital, Baltimore, MD 
21287 USA}
\tfootnote{This work was supported in part with funding from the NIH NIBIB Trailblazer Award (R21 EB024700). \\ Preprint - Under Review \copyright 2020 IEEE }


\corresp{Corresponding author: Nicholas J. Durr (e-mail: ndurr@jhu.edu).}

\begin{abstract}
While data-driven approaches excel at many image analysis tasks, the performance of these approaches is often limited by a shortage of annotated data available for training. Recent work in semi-supervised learning has shown that meaningful representations of images can be obtained from training with large quantities of unlabeled data, and that these representations can improve the performance of supervised tasks. Here, we demonstrate that an unsupervised jigsaw learning task, in combination with supervised training, results in up to a 9.8\% improvement in correctly classifying lesions in colonoscopy images when compared to a fully-supervised baseline. We additionally benchmark improvements in domain adaptation and out-of-distribution detection, and demonstrate that semi-supervised learning outperforms supervised learning in both cases. In colonoscopy applications, these metrics are important given the skill required for endoscopic assessment of lesions, the wide variety of endoscopy systems in use, and the homogeneity that is typical of labeled datasets. 
\end{abstract}



\maketitle

\section{Introduction}
\label{sec:introduction}
\PARstart{C}{olorectal} cancer is the second leading cause of cancer death and will cause a predicted 53,200 deaths in the United States in 2020 \cite{siegel2020}. Optical colonoscopy is considered the gold-standard for detecting and preventing colorectal cancer with approximately 15 million procedures being performed annually \cite{joseph2016}. Screening procedures are used to inspect the large intestine and rectum for precancerous lesions so that they may be removed prior to the onset of carcinoma. These lesions come in a variety of geometries and textures, each with an associated risk of progressing to a cancerous state \cite{colucci2003}. Colonoscopists analyze optical images to visually classify lesions, using cues such as color, shape, and vasculature patterns in conjunction with published guidelines \cite{hayashi2013, li2014, ijspeert2016}. Improving the reliability of lesion classification from images and de-skilling this task could reduce the costs, time, and other resources associated with histopathology. Further, lesions which are benign in nature may be left in place, eliminating associated risks of polyp removal \cite{anderloni2014}.

In the past decade, deep learning models have achieved astounding success in the computer vision field on tasks such as image classification and object recognition, surpassing human-level performance in some cases. In medical imaging, these models have outperformed traditional image processing techniques in a variety of fields such as radiology, histopathology, retinopathy, and  mammography. Most of these models are trained in a \textit{supervised} fashion, requiring large quantities of expertly annotated medical data to achieve optimal performance. In the medical imaging field, compiling annotated data is particularly time consuming, expensive, fraught with privacy concerns, and limited by the availability of expert annotators. In contrast, \textit{unsupervised} methods have shown that meaningful representations can be extracted from unlabeled data, which is often plentiful. In this work, we leverage the advantages of both labeled and unlabeled data using a \textit{semisupervised} learning paradigm to improve the performance of colonoscopy lesion classification.

Semi-supervised learning (SSL) is an emerging area of research that aims to learn a supervised objective, while enriching the encoded features through an unsupervised task. Recent works have shown marked improvement over purely supervised training, especially with small quantities of labeled data \cite{he2019, chen2020, caron2020}. SSL involves simultaneously training an unsupervised \textit{proxy} task, and a supervised task. Many proxy tasks involve applying some type of transformation to an image, then tasking the network with predicting the transformation. In this way, the network learns to encode information to a feature space which may enhance the performance of the supervised task. One example of a pretext task is applying a known rotation to an image, then tasking the network with estimating the degree of rotation.

In this paper, we use a jigsaw puzzle as the proxy task for SSL, as was first proposed by \cite{noroozi2016}. In this task, an input image is cut into an $N\times N$ grid, and the resulting tiles are reshuffled into an order defined by a randomly selected pseudo-label. The network then learns to encode the shuffled image into a feature vector which allows it to accurately predict the tile order. The unsupervised jigsaw task ideally enriches the encoder's resultant feature vectors, making them more discriminative for the supervised lesion classification task. Using this method, we find that a semi-supervised learning model outperforms a purely supervised model in lesion classification. While most semi-supervised learning research focuses solely on improvements in accuracy, trained models also benefit from improved robustness and generalizability. We also investigate the jigsaw method's effect on domain adaptation and out-of-distribution detection in colonoscopy - important metrics when deploying models to real-world clinical settings. Specifically, the contributions of this study are:
\begin{enumerate}
    \item To the best of our knowledge, this is first research applying semi-supervised learning to colonoscopy lesion classification.
    \item We demonstrate that a jigsaw-puzzle-solving task can effectively leverage unlabeled data to significantly improve the performance of lesion classification.
    \item We show that semi-supervised learning also improves performance in analyzing domain-shifted images and detecting out-of-distribution samples at inference.
\end{enumerate}

\section{BACKGROUND \& previous work}
\subsection{LESION CLASSIFICATION}
Polyp classification is a widely researched problem in the medical image analysis community \cite{min2019overview, nogueira2020deep}. Previous work has used traditional methods for hand-crafted feature extraction using color, texture, and 3D features for polyp classification in videos \cite{mesejo2016computer}. More recent research uses deep learning models, which have shown significant improvements in classification accuracy. Most use transfer learning \cite{zhang2016automatic} with off-the-shelf models such as ResNet \cite{lui2019endoscopic} and Inception \cite{byrne2019real, chen2018accurate, kandel2019su1741}. Others have combined traditional methods with deep learning approaches, such as fused wavelets and convolutional neural network features \cite{billah2017automatic}. Multi-modal fusion of pixel-level information, such as color and depth, have also been shown to improve classification accuracy \cite{mahmood2018multimodal, mahmood2019polyp}. Still, all of these methods exclusively utilize data with ground truth annotations \cite{ahmad2019artificial}.

\subsection{SELF-SUPERVISED \& SEMI-SUPERVISED LEARNING}
Self-supervised and semi-supervised learning are highly active areas of artificial intelligence research. These methods exploit unlabeled data for effective representation learning. Recent semi-supervised works have achieved comparable performance to conventional fully supervised networks, while only requiring a small fraction of labeled data. To learn from data without manual annotations, self-supervised methods employ proxy tasks where pseudo-labels can be generated using know transformations or data manipulations. According to \cite{jing2019}, there are four common types of proxy tasks:
\begin{itemize}
  \item Generation-based methods: Some part of the data is deliberately removed, and the network is tasked with predicting the missing data. Examples include image colorization \cite{zhang2016automatic}, image inpainting \cite{pathak2016context}, and video generation from single frames using generative adversarial networks (GANs) \cite{vondrick2016generating}.
  \item Context based methods: The network is tasked with learning to make predictions using either spatial or temporal contextual information. Examples include image clustering \cite{caron2018deep, caron2020}, context prediction \cite{doersch2015unsupervised, noroozi2016, oord2018representation}, predicting a geometric transformation such as rotation \cite{gidaris2018unsupervised}.
  \item Free semantic label-based methods: Semantic labels are automatically generated for object segmentation \cite{pathak2017learning, croitoru2017unsupervised} or contour detection \cite{ren2018cross, li2016unsupervised, albuquerque2020improving}. 
  \item Cross modal methods: Data correspondence between data modalities is learned such as Visual-Audio Correspondence \cite{korbar2018cooperative, arandjelovic2017look}.
\end{itemize}

Recent works have shown that semi-supervised learning methods improve model robustness and generalizability, as well as the ability to measure uncertainty \cite{hendrycks2019, carlucci2019}. Deep learning models are notorious for silently providing incorrect predictions when test samples are drawn from a distribution other than the distribution used for training. Surrogate methods have been incorporated into the inference pipeline, drawing on the network's prediction probabilities to determine an out-of-distribution score for test samples \cite{hendrycks2016}. The success of semi-supervised learning in medical imaging is dependent on deploying networks that can handle a wide distributing of samples, and have a mechanism for appropriately handling samples which the network is ill-conditioned to classify.

\begin{figure*}[h]
    \centering
    \includegraphics[width=7in]{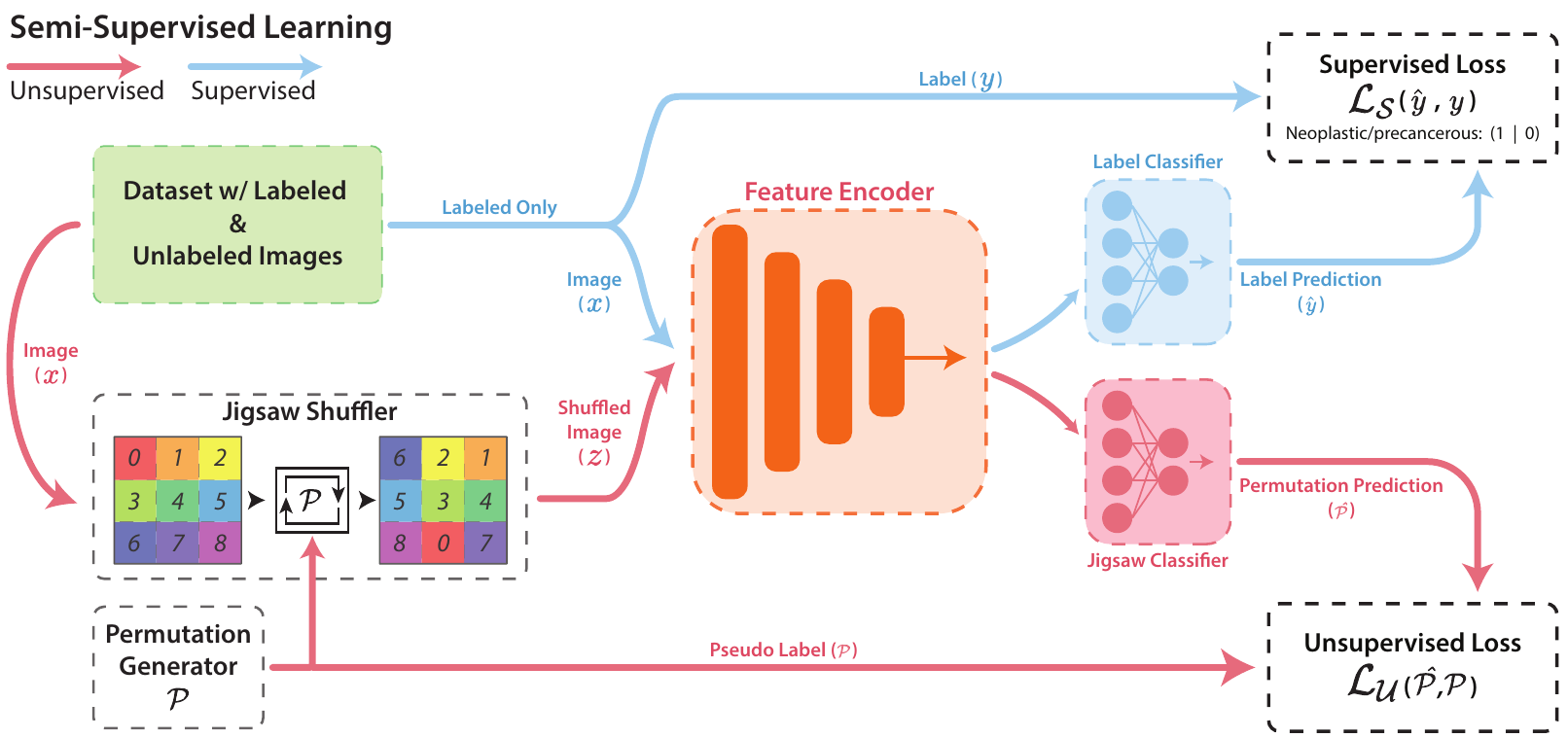}
    \caption{The proposed semi-supervised learning model uses lesion type labels for a supervised loss and jigsaw index pseudo labels for an unsupervised loss. This model is sequentially trained in a supervised phase then an unsupervised phase for each iteration.}
    \label{fig:overview}
\end{figure*}

\subsubsection{SEMI-SUPERVISED LEARNING IN MEDICAL IMAGING}
Since labeled data in medical imaging community is particularly scarce, researchers in this field have long explored unsupervised methods. Cheplygina et al. \cite{cheplygina2019not} present a comprehensive review of semi-supervised and self-supervised methods employed in medical imaging. Popular approaches include using self-labeling and co-training, where a classifier is first trained on the available labeled data, and is then used to generate pseudo labels on unlabeled data. The classifier is then retrained using the newly generated labeled data. This method is especially popular where precise labeling is cumbersome, such as pixel-level segmentation tasks with applications in neuro \cite{iglesias2010agreement, meier2014patient, dittrich2014spatio}, heart \cite{gu2017semi}, and retinal \cite{bai2017semi} imaging.
More recent works have employed state-of-art semi-supervised and self-supervised techniques across a wide range of applications, such as consistency regularization for skin lesion classification and thorax disease diagnosis \cite{wang_focalmix:_2020}, unsupervised anomaly detection for white matter lesion segmentation \cite{baur2019fusing}, and image synthesis with GANs for data augmentation in glaucoma assessment \cite{diaz-pinto_retinal_2019}.

\subsubsection{JIGSAW PUZZLE SOLVING} 
The original semi-supervised jigsaw approach proposes decomposing an image into patches, shuffling the patches, then individually feeding the patches to a Siamese network \cite{noroozi2016}. The network predicts the shuffled patch order as a pretext task, and it is later fine-tuned on the downstream, supervised task using labeled data. Many variations of the jigsaw task have been explored, including for videos \cite{ahsan_video_2019}, three-dimensional data \cite{zhuang2019self}, and negative sample inclusion for increased difficulty \cite{noroozi_boosting_2018}. Specifically in medical imaging, the jigsaw paradigm has been applied to imaging of the brain and pancreas \cite{taleb2019multimodal, zhuang2019self, tao2020revisiting}.

In this work, we adapt the jigsaw proxy task for improving the performance of a supervised classifier \cite{carlucci2019}. To the best of our knowledge, this work is the first to explore semi-supervised learning for lesion classification in colonoscopy. The most similar prior art is \cite{ross2018exploiting} which performs medical instrument segmentation on endoscopy images using image colorization as the pretext task.

\section{METHODS}

Our problem statement is defined as follows: given a colonoscopy image of a lesion, we attempt to classify it into one of two classes - neoplastic/precancerous or non-neoplastic. Our dataset consists of labeled and unlabeled image sets, $\mathcal{D} = \mathcal{D}_{\mathcal{L}} \cup \mathcal{D}_{\mathcal{U}}$, where $\mathcal{D}_{\mathcal{L}}$ consists of image-label pairs $\mathcal{D}_{\mathcal{L}}=\left\{x_{l}^{i}, y_{l}^{i}\right\}_{i=1}^{N_{l}}$ with $N_{l}$ as the total number of labeled images, and $\mathcal{D}_{\mathcal{U}}$ is the set of unlabeled lesion images, $\mathcal{D}_{\mathcal{U}}=\left\{x_{u}^{i}, \right\}_{i=1}^{N_{u}}$ where $N_{u}$ is the total number of unlabeled images. Detailed description of the classes \& dataset is given in section \ref{sec:dataset}. The goal is to leverage the unlabeled data $\mathcal{D}_{\mathcal{U}}$ using the jigsaw task to improve the performance of lesion classification.

\subsection{ARCHITECTURE}
\label{sec:archi}
As shown in Figure \ref{fig:overview}, our model consists of ResNet-18 as a shared feature encoder with two classifier heads - one for supervised lesion classification and a second for jigsaw classification. Our deep model is denoted by $f$, where the shared feature extractor is parameterized by $\theta_{e}$ and the supervised and unsupervised classifier heads by $\theta_{s}$ and $\theta_{u}$, respectively. The network trains in two phases - a supervised phase that minimizes the supervised loss $\mathcal{L}_{\mathcal{S}}$ followed by an unsupervised phase that minimizes the jigsaw loss $\mathcal{L}_{\mathcal{U}}$. The parameters of the network are learned by alternating training between the supervised and unsupervised tasks on each iteration. The following sections describe the two training phases in detail.

\subsubsection{SUPERVISED PHASE}
The main supervised objective is to classify colonoscopy lesion images into neoplastic vs non-neoplastic classes. We aim to minimize the supervised classification loss $\mathcal{L}_{\mathcal{S}}$, which is the weighted cross-entropy loss between the target label $y_i$ and the model prediction $f(x_i|\theta_{f},\theta_{s})$ with $(x_i, y_i) \in \mathcal{D}_{\mathcal{K}}$. In our experiments to assess the effectiveness of semi-supervised learning, we report the performance of the network trained on various fractions of the labeled dataset. Consequently, $\mathcal{D}_{\mathcal{K}} \subseteq \mathcal{D}_{\mathcal{L}}$ is obtained by selecting the $k^{th}$ percentage of labeled data where $k$ varies logarithmically i.e. $k = \{100, 50, 25, 12.5, 6.25\}$. A detailed description of how data selection is performed is discussed in \ref{sec:vld}. The cross entropy loss function is weighted to account for the class imbalance in the dataset. Formally, the supervised loss function is defined as 
\begin{equation}
    \mathcal{L}_{\mathcal{S}} = -\frac{1}{|\mathcal{D}_{\mathcal{K}}|}\Sigma_{i=1}^{|\mathcal{D}_{\mathcal{K}}|} \Sigma_{c=0}^{1} w_c y_{i,c} log(p(y_{i,c}|x_i, \theta_{e}, \theta_{s})),
    \label{equ:sup_loss}
\end{equation}
where $|\mathcal{D}_{\mathcal{K}}|$ is the number of images in the selected labeled dataset, weight $w_c = 1/freq(c)$ is the inverse class frequency $c$ in the dataset $\mathcal{D}_{\mathcal{K}}$, $y_{i,c}$ is the one-hot encoded target label for the $i^{th}$ image, and $p$ is the posterior probability obtained by taking the softmax of output logits from $f$. In this phase, only parameters of the feature encoder $\theta_{e}$ and supervised fully connected layer $\theta_{s}$ are updated. 

\subsubsection{UNSUPERVISED PHASE}
\begin{figure}
    \centering
    \includegraphics[width=\linewidth]{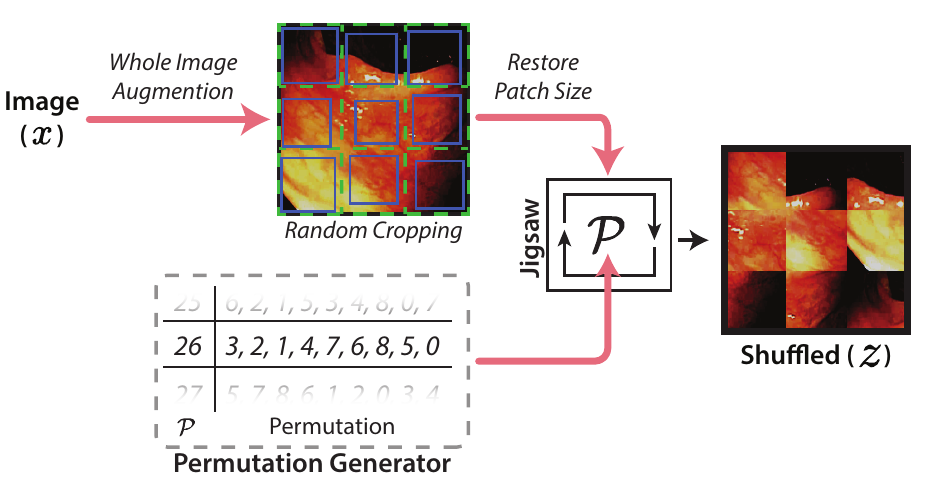}
    \caption{Overview of the jigsaw shuffler procedure for generating shuffled images with a pseudolabel for unsupervised learning.}
    \label{fig:jig_shuffler}
\end{figure}

Following each supervised phase, an unsupervised phase is trained using the entire dataset $\mathcal{D}$. In this phase, the objective of the network is to learn to solve the jigsaw task. As shown in Figure \ref{fig:jig_shuffler}, we first decompose an image into a $3 \times 3$ grid of tiles. Then, a patch of 0.75-0.9 times the original tile size and a random offset is cropped from each tile. The patches are then scaled back to the original tile size, reordered according to a selected permutation index $\mathcal{P}$, and concatenated to reform a $222\times222$ input image $z$. This transformation prevents the network from using low level cues such as continuity of edges, color, or texture when estimating the patch order. Instead, the network is forced to learn high-level, global primitives such as shape. With $9$ grid positions, there are $9!$ possible patch permutations, creating far too many labels for the network to learn. To make the classification task achievable for the network, we select a small subset of the possible $P$ permutations with maximal Hamming distance from one another \cite{noroozi2016}. An index is assigned to each permutation, which then functions as a pseudo-label. The jigsaw task is then formulated as a classification problem, tasking the network to learn to correctly predict the pseudo label $\mathcal{P} \in \{0, 1, 2, ..., P\}$ of $z$. Here, the zero index refers to the unscrambled, original image case.

We use a weighted cross-entropy loss as the unsupervised loss $\mathcal{L}_{\mathcal{U}}$.
When creating a mini-batch for training in the unsupervised phase, we keep the scrambled-to-unscrambled image ratio equal to $s:(1-s)$, where $s\in[0,1]$. In the jigsaw shuffler, the permutation index for the scrambled images is drawn from uniform distribution $\mathcal{U}\{1, P\}$. Hence, the frequency of occurrence for permutation indices is $freq = ((1-s), s/P, s/P, ..., s/P)$, where $freq(\mathcal{P})$ is the frequency of permutation index $\mathcal{P}$. The inverse of frequency is used as a scalar weighting in the cross entropy loss, $w_{\mathcal{P}} = 1/freq(\mathcal{P})$. The unsupervised loss is defined as follows :
\begin{equation}
    \mathcal{L}_{\mathcal{U}} = -\frac{1}{|\mathcal{D}|}\Sigma_{i=1}^{|\mathcal{D}|} \Sigma_{\mathcal{P}=0}^{P} w_{\mathcal{P}} y_{i,\mathcal{P}} log(p(y_{i,\mathcal{P}}|z_i, \theta_{e}, \theta_{u})) 
    \label{equ:unsup_loss}
\end{equation}
where $|\mathcal{D}|$ is the total number of images in the training dataset, $z_i$ is the $i^{th}$ recomposed image, $y_{i, p}$ is the one hot encoded pseudo label vector, and $p(y_{i,p}|z_i, \theta_{e}, \theta_{u})$ is the prediction probability for the $p^{th}$ permutation. Minimization of the unsupervised loss involves only learning the feature encoder $\theta_{e}$ and the unsupervised head $\theta_{u}$. 

The overall training loss $\mathcal{L}_{total}$  is then : 
\begin{equation}
    \mathcal{L}_{total} = \mathcal{L}_{\mathcal{S}} + \lambda \mathcal{L}_{\mathcal{U}}
\end{equation}
where $\lambda$ is a scalar weight applied to the unsupervised loss. In the unsupervised phase, ordered and shuffled images are mixed. During the supervised phase, input images remained ordered, just as they are presented during testing. When training is complete, the unsupervised head is discarded, and only the trained feature encoder and supervised lesion classification head are used for testing.

\subsection{DOMAIN ADAPTATION}
This section describes experiments to assess how semi-supervised learning impacts the domain generalizability of a model. In the context of colonoscopy, domain adaption would be useful when applying a network to new endoscope types or manufacturers, to endoscopes with imaging performance that varies over time (e.g. dirty optics), or to new imaging modes. We experimentally withhold a target domain of data from the supervised task and only include it in the unlabeled set for the unsupervised task. We can then assess the domain adaptability of the network by testing on labeled samples from the target domain.

In colonoscopy, two widely used imaging modalities are White Light Imaging (WLI), and Narrow Band Imaging (NBI). For our experiment, we consider WLI as the source domain, and NBI as the target domain. For training, we use labeled WLI image-lesion class label pairs, $\mathcal{D}_{\mathcal{L-WLI}} = \{x_i, y_i\}_{i=0}^{N_{L-WLI}}$ where $\mathcal{D}_{\mathcal{L-WLI}} \subset \mathcal{D}_{\mathcal{L}}$ and $N_{L-WLI}$ is the total number of white light labeled images. We also use unlabeled NBI images $\mathcal{D}_{\mathcal{U-NBI}}=\{x_i\}_{i=0}^{N_{U-NBI}}$ where $N_{U-NBI}$ is the total number of unlabeled NBI images and $\mathcal{D}_{\mathcal{U-NBI}} \subset \mathcal{D}_\mathcal{U}$. For testing the performance of the network, we use labeled NBI images, $\mathcal{D}_{\mathcal{L-NBI}} = \{x_i, y_i\}_{i=0}^{N_{L-NBI}}$ where $N_{L-NBI}$ is the total number of labeled NBI images and $\mathcal{D}_{\mathcal{L-NBI}} \subset \mathcal{D}_{\mathcal{L}}$.

The network training approach remains the same as was described in the previous section, with the only exception being the data used in each phase. In the supervised phase, we use the labeled WLI images from $\mathcal{D}_{\mathcal{L-WLI}}$, whereas in the unsupervised phase we use both the labeled WLI images and the unlabeled NBI images i.e. $\mathcal{D}_{\mathcal{L-WLI}} \cup \mathcal{D}_{\mathcal{U-NBI}}$. In the testing phase, we use labeled NBI images $\mathcal{D}_{\mathcal{L-NBI}}$.

\begin{figure}
  \centering
  \begin{subfigure}[b]{0.49\linewidth}
  \centering
    \includegraphics[width=\textwidth]{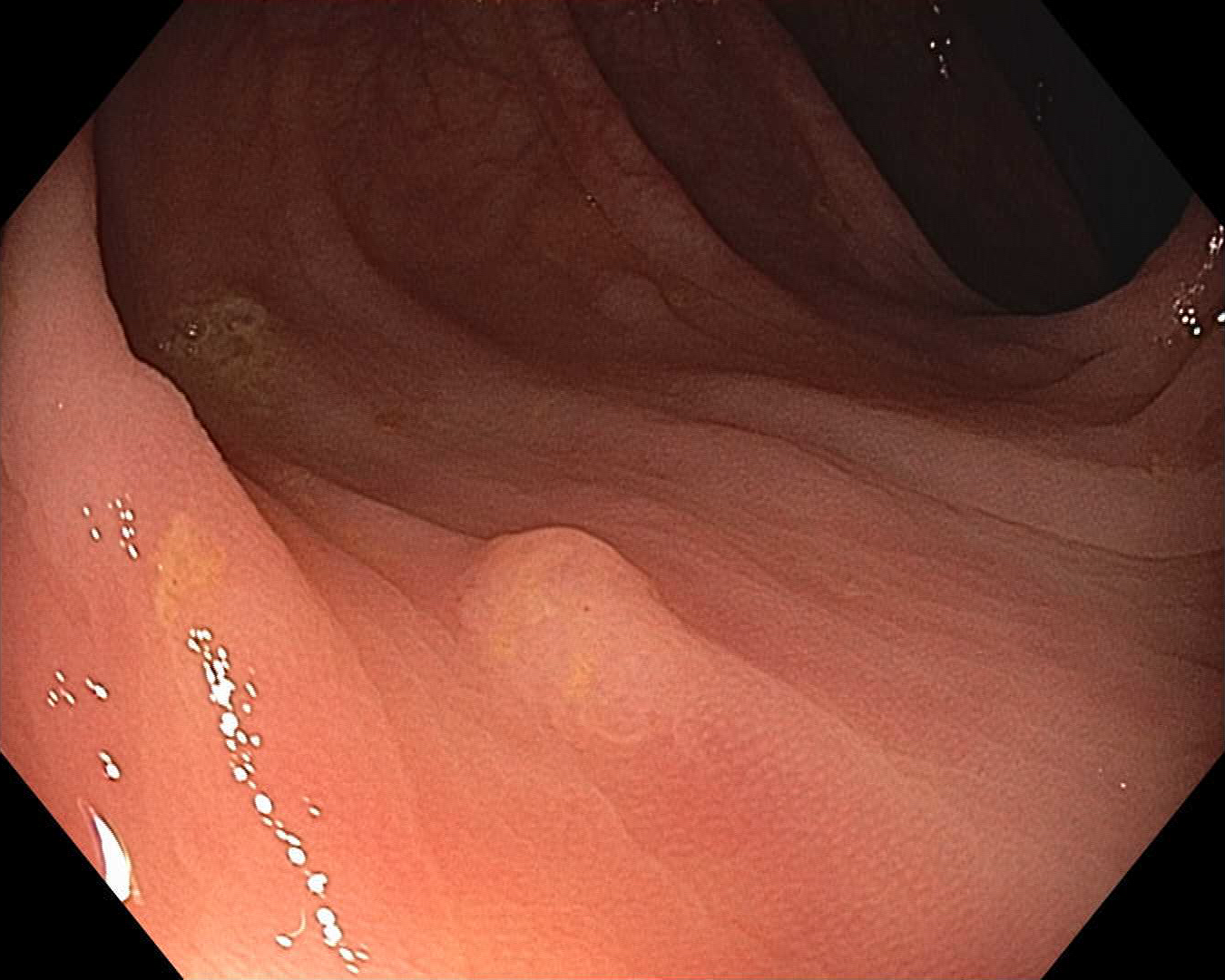}
    \caption{White Light Imaging (WLI)}
  \end{subfigure}
  \hfill
  \begin{subfigure}[b]{0.49\linewidth}
  \centering
    \includegraphics[width=\textwidth]{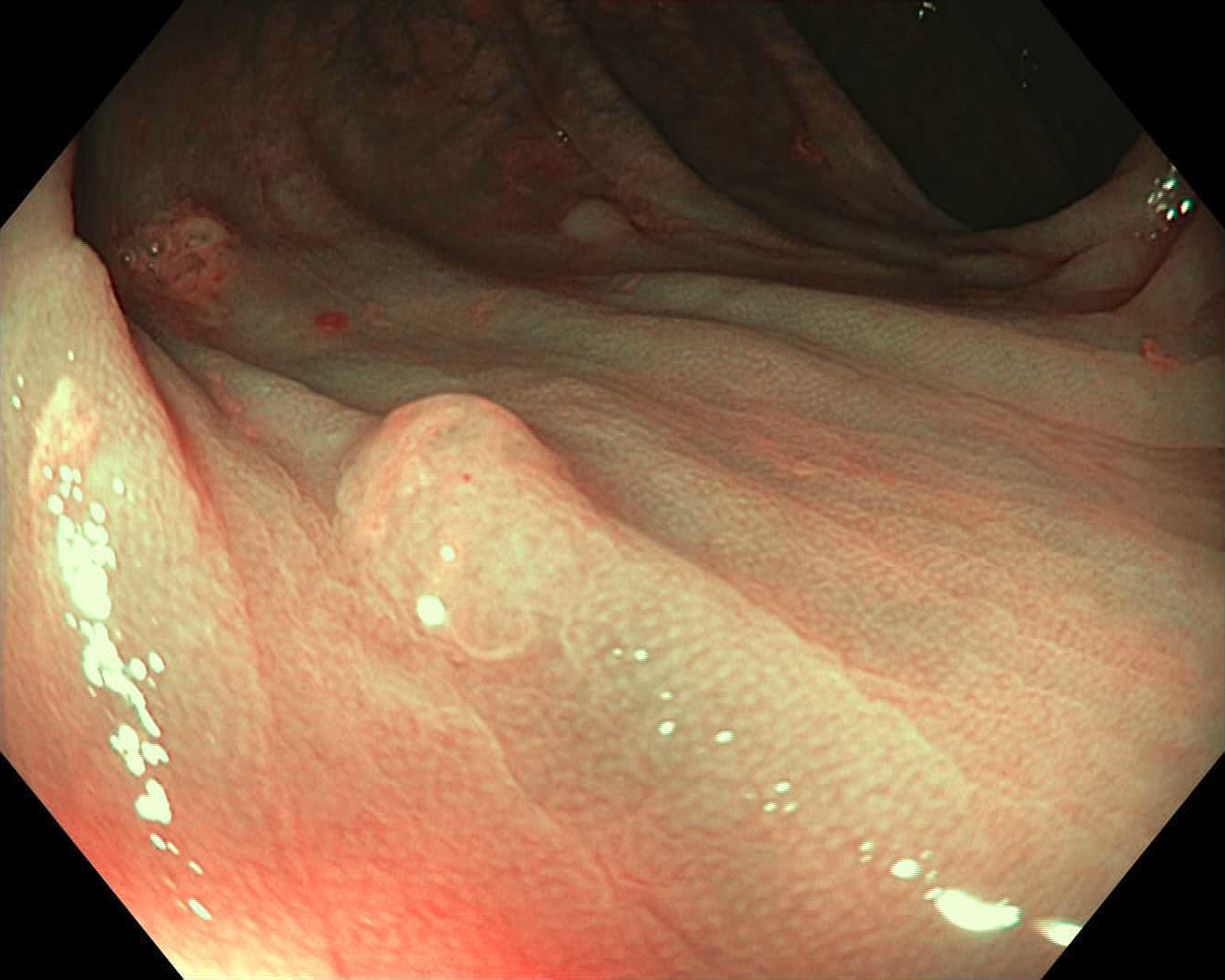}
    \caption{Narrow Band Imaging (NBI)}
  \end{subfigure}
  
  \caption{Illustrative colonoscopy polyp images showing difference in the WLI \& NBI modalities}
  \label{fig:wli_nbi}
\end{figure}

\subsection{OUT-OF-DISTRIBUTION DETECTION}
In out-of-distribution detection, the goal is to identify test samples which don't belong to the distribution on which the model was trained. These out-of-distribution samples can then be rejected to avoid unreliable inference. A pretrained semi-supervised learning model can act as an efficient out-of-distribution detector. In this experiment, we train a classifier using in-distribution samples on the main objective of lesion classification, and then later test its performance as an out-of-distribution detector. We consider white light images to be in-distribution samples, and NBI images are treated as out-of-distribution samples. In the supervised phase, we use labeled white light images from $\mathcal{D}_{\mathcal{L-WLI}}$. For the unsupervised phase, we use unlabeled and labeled white image i.e. $\mathcal{D}_{\mathcal{L-WLI}} \cup \mathcal{D}_{\mathcal{U-WLI}}$. To use the classifier as an out-of-distribution detector, we utilize the posterior probabilities $p(y|x)$. Is is shown in \cite{hendrycks2016baseline, hendrycks2018deep} that the probability distribution of prediction softmax probabilities for out-of-distribution samples appears roughly uniform in distribution. Whereas, in-of-distribution samples have a more 'peaky' distribution with a higher maximum softmax probability $max_c p(y=c|x)$. An out-of-distribution detector score $\kappa$ based on the 
posterior probabilities and the auxiliary jigsaw loss is defined as follows:
\begin{equation}
    \kappa = KL[U||p(y|x)] - \Sigma_{\mathcal{P}=0}^{P} w_{\mathcal{P}} y_{i,\mathcal{P}} log(p(y_{i,\mathcal{P}}|z_i, \theta_{e}, \theta_{u}) 
    \label{equ:ood_score}
\end{equation}
where $KL[U||p(y|x)]$ is the KL-divergence between the uniform distribution and the prediction softmax probabilities and $\Sigma_{\mathcal{P}=0}^{P} w_{\mathcal{P}} y_{i,\mathcal{P}} log(p(y_{i,\mathcal{P}}|z_i, \theta_{e}, \theta_{u})$ is the unsupervised loss for image $x$ as defined in Equation \ref{equ:unsup_loss}. KL divergence measures the difference between two probability distributions. If two probability distributions are similar, the KL divergence between them is low, whereas a high value indicates that they are starkly different. The KL divergence between distributions $P(y)$ \& $Q(y)$ is defined as :
\begin{equation}
    {KL}[Q \| P]=\sum_{y} Q(y) \log \left(\frac{Q(y)}{P(y)}\right)
\end{equation}
where $y$ is the support of the distribution i.e $y \in \{0, 1\}$ for this case. In the baseline experiment, the OOD score is $\kappa = KL[U||p(y|x)]$. For the semi-supervised learning case, we also add the jigsaw cross entropy loss. For testing, we use unseen WLI images as the negative class (label = 0) and NBI images as positive class (label = 1). The in-distribution trained polyp classifier is used for inference of the test set to generate the OOD score $\kappa$. It is important to note that training the classifier doesn't have any element of OOD, and it is trained solely to classify lesions. Another advantage is that this approach doesn't require any OOD samples during training. Distinguishing WLI \& NBI images by itself is not a clinically motivated problem, but we use it as a proxy setup to demonstrate SSL's potential as an OOD detector. 

\section{EXPERIMENTS \& RESULTS}
\subsection{DATASET}
\label{sec:dataset}
\begin{table}[]
\caption{\textbf{Summary of dataset}}
\label{tab:dataset}
\resizebox{\linewidth}{!}{%
\begin{tabular}{|l|l|l|l|}
\hline
\multicolumn{2}{|l|}{Total Number of frames} & \multicolumn{2}{c|}{6,649} \\ \hline
Labeled frames             & 4,095            & Unlabled frames   & 2,554  \\ \hline
Neoplastic Frames              & 3,369            & Non-Neoplastic frames  & 726   \\ \hline
WLI Frames                 & 3,855            & NBI Frames        & 2,646  \\ \hline
\end{tabular}%
}
\end{table}

The colonoscopy video data used in this paper was collected at the Johns Hopkins Hospital using a protocol approved by the Johns Hopkins Institutional Review Board (\#IRB00184221). Video segments were analyzed and cropped from patient procedure video data, retrospectively, to limit included frames to those containing lesions that were biopsied by the endoscopist. Tissue biopsies were collected from suspected lesions, and ground truth labels derived from histopathology analysis were later paired with the respective video segments. A total of 108 patients were enrolled in the study. A total of 132 videos with corresponding ground truth labels were collected, with each video segment featuring a unique lesion. Video annotations were recorded by two medical trainees and verified by an experienced gastroenterologist. An additional 112 videos with no ground truth classification were cropped and extracted for training the semi-supervised model.

Videos were further categorized into two classes: "neoplastic/precancerous" and "non-neoplastic". Using the histologic labels, adenomas and serrated adenomas were assigned to the neoplastic/precancerous class (n=110), while hyperplastic polyps were assigned to the non-neoplastic class (n=22). The videos include a diverse distribution of imaging parameters, such as varied video processors, illumination modes (WLI/NBI), as well as scope manufacturer and models with high- \& standard-definition resolutions. Videos were separated into training and testing sets with equal class balance between sets. Derived image frames were stored in separate containers to prevent class leakage. Repetitive image frames resulting from minimal camera motion were discarded. A frame wise summary of the dataset is given in table \ref{tab:dataset}.

\subsection{IMPLEMENTATION DETAILS}

\begin{table*}[]
\caption{\textbf{Descriptive statistics comparing the performance of  semi-supervised against baseline as function of labeled data percentage. The median values across 5-fold cross-validation are reported.}}
\label{tab:expt1}
\resizebox{\textwidth}{!}{%
\begin{tabular}{c|cccccccccc}
\hline\hline
\multirow{2}{*}{Labeled Data } & \multicolumn{2}{c}{Accuracy (\%)}                           & \multicolumn{2}{c}{F1 Score}                           & \multicolumn{2}{c}{Sensitivity}                        & \multicolumn{2}{c}{Specificity}                        & \multicolumn{2}{c}{Precision}                          \\ \cline{2-11} 
                                 & \multicolumn{1}{l}{Baseline} & \multicolumn{1}{l}{SSL} & \multicolumn{1}{l}{Baseline} & \multicolumn{1}{l}{SSL} & \multicolumn{1}{l}{Baseline} & \multicolumn{1}{l}{SSL} & \multicolumn{1}{l}{Baseline} & \multicolumn{1}{l}{SSL} & \multicolumn{1}{l}{Baseline} & \multicolumn{1}{l}{SSL} \\ \hline
6.25\%                           & 57.04                        & \textbf{66.80}          & 0.69                         & \textbf{0.79}           & 0.55                         & \textbf{0.68}           & \textbf{0.70}                & 0.52                    & \textbf{0.94}                & 0.89                    \\
12.5\%                           & 60.53                        & \textbf{69.67}          & 0.73                         & \textbf{0.79}           & 0.63                         & \textbf{0.76}           & \textbf{0.38}                & 0.22                    & \textbf{0.87}                & 0.83                    \\
25\%                             & 68.86                        & \textbf{71.96}          & 0.77                         & \textbf{0.81}           & 0.68                         & \textbf{0.80}           & \textbf{0.33}                & 0.19                    & \textbf{0.89}                & 0.82                    \\
50\%                             & 69.54                        & \textbf{75.60}          & 0.80                         & \textbf{0.85}           & 0.73                         & \textbf{0.88}           & \textbf{0.54}                & 0.20                    & \textbf{0.85}                & 0.80                    \\
100\%                            & 73.91                        & \textbf{76.76}          & 0.82                         & \textbf{0.85}           & 0.83                         & \textbf{0.87}           & \textbf{0.37}                & 0.24                    & \textbf{0.82}                & 0.82                    \\ \hline\hline
\end{tabular}%
}
\end{table*}

All experiments are implemented using PyTorch library \cite{NEURIPS2019_9015} on a server equipped with an NVIDIA RTX 2080Ti 11GB GPU, an Intel Xeon Processor W-2123 3.6 GHz CPU, and 64 GB of RAM. We use the JiGen repository \cite{carlucci2019} as our base code for development. All experiments utilize ResNet-18 \cite{he2016} as the feature encoder. The fully connected layers are $512\times2$ for the supervised branch and $512\times P$ for the unsupervised branch, similar to the FCN classifier in ResNet-18, only differing by the number of output nodes. 

The network weights are initialized using the pre-trained ImageNet ResNet-18 weights available in the PyTorch library. Data augmentation for whole images includes random vertical flip, random horizontal flip, random rotation into \{$0^{\circ}, 90^{\circ}, 180^{\circ}, 270^{\circ}$\}, and random crops of size [0.8, 1.0] (all p=0.5). The images are normalized with mean [0.485, 0.456, 0.406] and standard deviation [0.229, 0.224, 0.225]. The augmented images are finally resized to $222\times222$. In the case of the unsupervised phase, the whole image transformations are applied before the jigsaw shuffler. No color transformations are applied, as polyp color is a discriminative feature among the classes. ADAM optimizer \cite{kingma2014method} with weight decay ($L_2$ Penalty) is used for training the network. The initial learning rate is kept as 0.0001. The ratio of frame-wise frequency of class is 0.83:0.17 for the neoplastic to non-neoplastic classes, the inverse of which is used as weights in the supervised weighted cross entropy in equation \ref{equ:sup_loss}. The scrambled to unscrambled image ratio $s:1-s$ used in equation \ref{equ:unsup_loss} is kept as 0.6:0.4.

\subsection{VARYING THE QUANTITY OF LABELED DATA}
\label{sec:vld}
To test the efficacy of our semi-supervised learning approach, we evaluate its performance as a function of the quantity of labeled data $\mathcal{D}_{\mathcal{K}}$ used for training. We train the network using $k\%$ of the total labeled training data where $k$ varied logarithmically, $k = \{100, 50, 25, 12.5, 6.25\}$. For each $k$, we perform a five-fold cross validation. To split the dataset, we first select 20\% of the total labeled data for validation. This split of validation set is done at the video level to prevent images of the same polyp mixing between the train and validation sets. Next, we choose $k\%$ of the remaining labeled datasets as our supervised training dataset $\mathcal{D}_{\mathcal{K}}$. Thus, the validation dataset for a particular fold remains the same for all the values of $k$. We use the selected labeled dataset $\mathcal{D}_{\mathcal{K}}$ for training the supervised phase, but for all values of $k$ we use the whole training dataset $\mathcal{D}$ (excludes the validation images) for the unsupervised phase. On an average there are 819 images in the validation set.

We perform an ablation study to measure the performance of SSL when compared to a baseline model. The baseline model is also a ResNet-18, and it is architecturally the same as the SSL model (described in \ref{sec:archi}), but without the jigsaw head. The baseline model uses the same weighted cross entropy loss that the SSL model uses in the supervised phase (Equation \ref{equ:sup_loss}). When comparing the performance of the SSL model and the baseline, both models use the same validation data and the selected labeled data for supervised training $\mathcal{D}_{\mathcal{K}}$.

The hyperparameters which gave the best performance for both models are reported. For the baseline model, an initial learning rate of 0.0001 is used for all cases except for $100\%$ models where 0.001 is used. As for weight decay, $100\%$ model uses 0.005, $50\%$ \& $25\%$ uses 0.05, $12.5\%$ has 0.2 and $6.25\%$ uses a value of 0.005. For the SSL models, an initial learning rate of 0.0001 is used. The number of jigsaw classes (P) is 30 for all cases except $100\%$, which uses 100 classes. The weight decay values are - 0.005 for $100\%$, 0.05 for $50\%$, 0.07 for $25\%$ \& $12.5\%$, and 0.2 for $6.25\%$. The unsupervised loss weights $\lambda$ are 1 for $100\%$ \& $50\%$, 2 for $25\%$, and 1.5 for $12.5\%$ \& $6.25\%$. The $\lambda$ value is also increased 1.5 times every 5 epochs for low data regime training to accelerate the unsupervised learning phase to match the swift learning on the supervised end, due to small labeled data size.

We evaluate the classification performance with five commonly used metrics - accuracy, F1 score, sensitivity, specificity and precision. Accuracy is the ratio of correct predictions over the total number of test samples. Since our data has an uneven class distribution, we also use F1 score for evaluation. F1 score is the harmonic mean of precision and recall. Sensitivity is the ratio of correctly predicted positive samples to the total number of positive samples (neoplastic/precancerous class). Similarly, specificity is the ratio of correctly classified negative class samples (non-neoplastic class). Precision is the ratio of correctly predicted positives to all predicted positives. Definitions are as follows: 
\begin{equation}
    \text{Accuracy} = \frac{TP+TN}{TP+TN+FP+FN}
\end{equation}
\begin{equation}
    \text{F1 Score} = \frac{2TP}{2TP+FP+FN}
\end{equation}
where true positive $TP$ is the number of correct predictions for the positive class while true negative $TN$ is the number of correct predictions for the negative class. False negative $FN$ is the number of samples incorrectly classified to negative class whereas false positives $FP$ is the incorrect classifications to the positive class. 

\begin{figure*}[h!]
  \centering
  
  \begin{subfigure}{0.47\textwidth}
    \includegraphics[width=\linewidth]{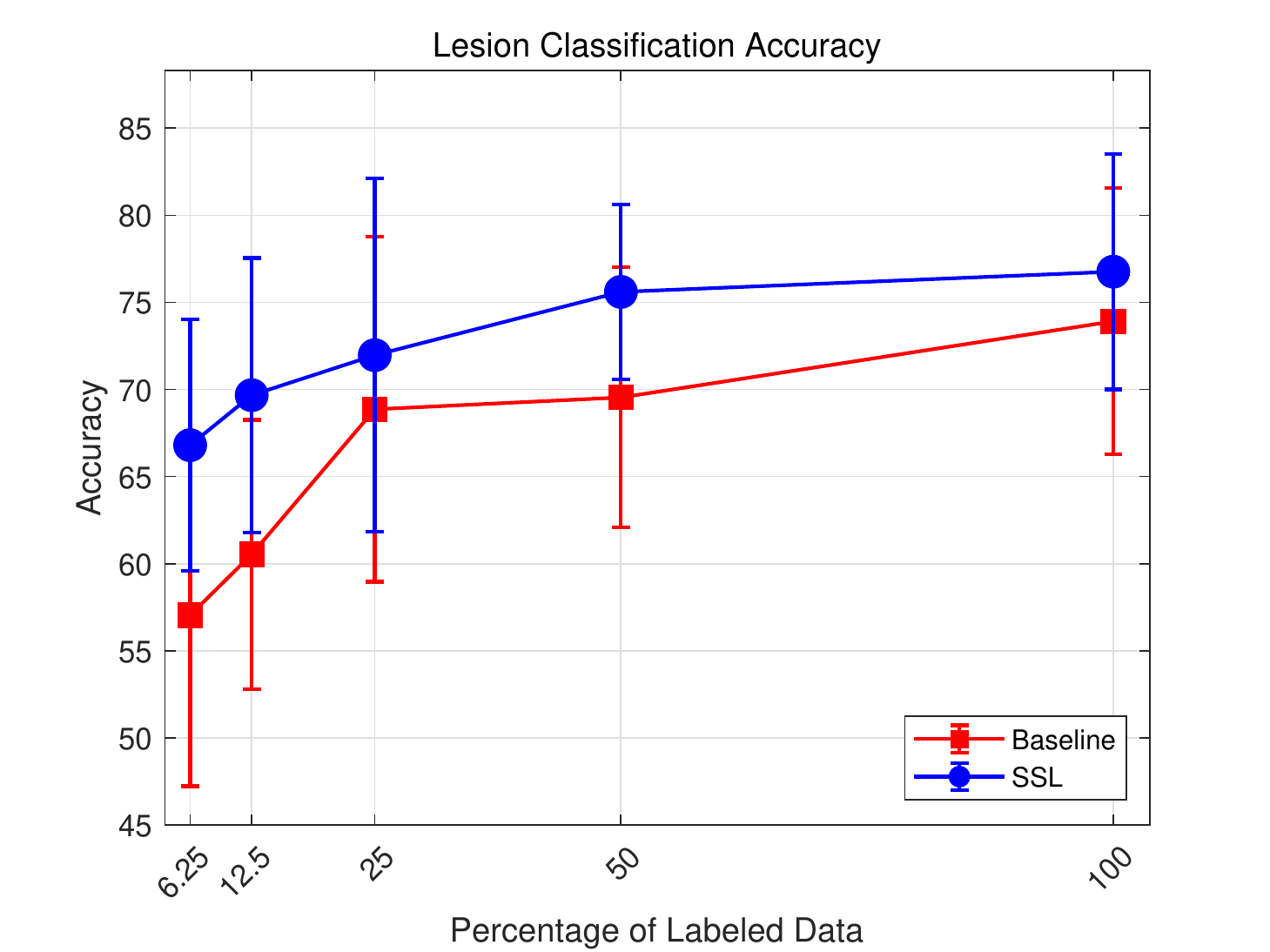}
  \end{subfigure}
  \begin{subfigure}{0.47\textwidth}
    \includegraphics[width=\linewidth]{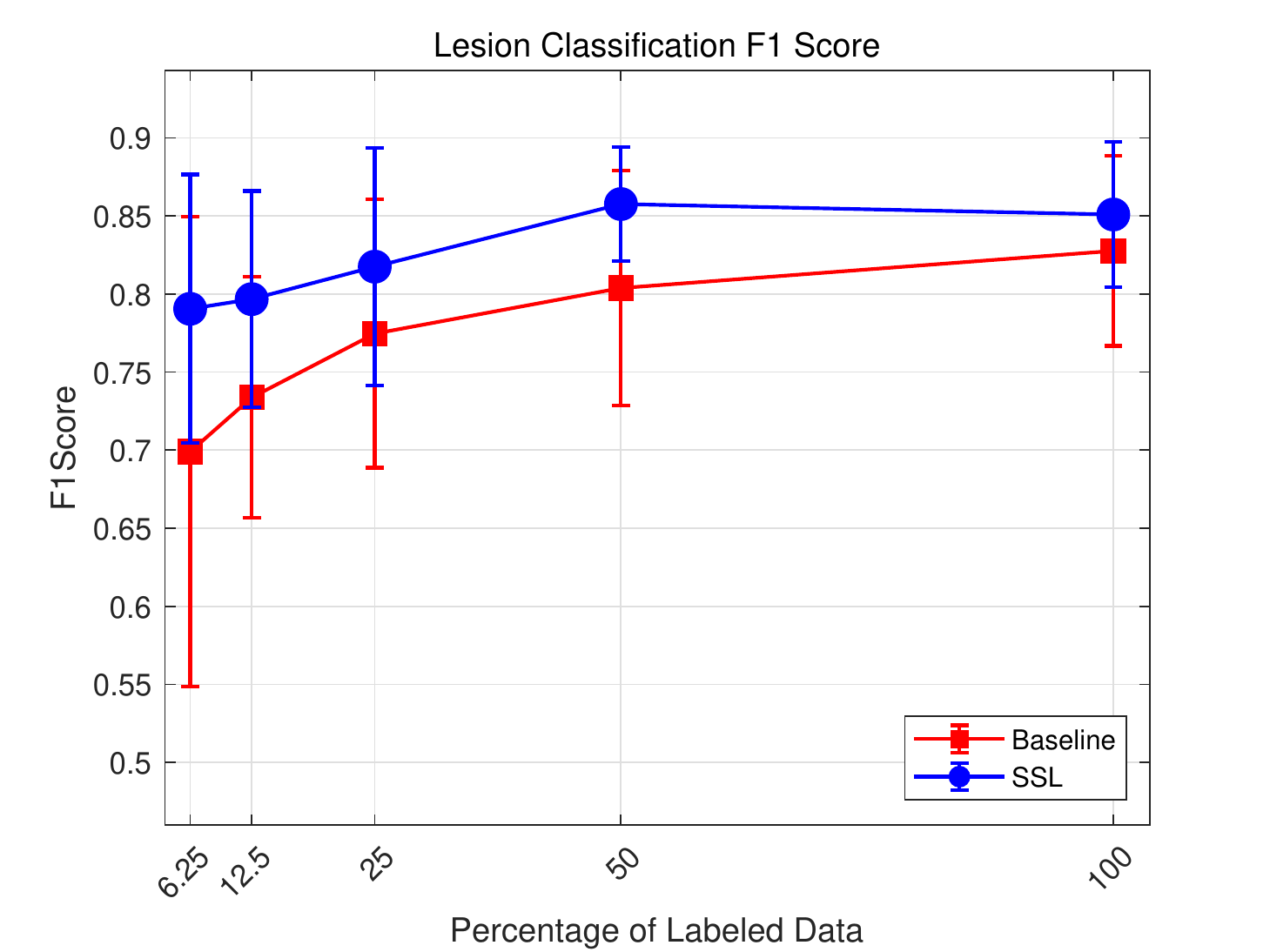}
    
  \end{subfigure}
  \caption{Results comparing semi-supervised learning against baseline as function of fraction of labeled training data. Median and standard deviation for 5-fold cross validation are reported.}
  \label{fig:expt1}
\end{figure*}

Figure \ref{fig:expt1} plots the median metrics and the standard deviation across the five fold cross validation as a function of percentage of labeled data. All the performance metrics are shown in Table \ref{tab:expt1}. From Figure \ref{fig:expt1}, we can observe that the semi-supervised learning consistently achieves superior performance compared to the baseline for all cases in terms of accuracy \& F1 Score. The accuracy for semi-supervised learning falls by only 9.96\% as compared to 16.87\% for baseline when moving to the low data regime. Similarly, for F1 score  we observe a drop of only 0.06 for semi-supervised learning versus a 0.13 drop by the baseline. With only 6.25\% labeled data, the semi-supervised model gives an accuracy of 66.80\% and a F1 score of 0.79. Using 100\% of the labeled data in conjunction with unlabeled data gave semi-supervised model a boost of 2.85\% in accuracy and 0.03 in F1 score. When comparing the sensitivity, the semi-supervised approach exceeds the baseline in all cases.

The semi-supervised improvement over the baseline indicates that adding a jigsaw solving auxiliary task is beneficial. This improvement could be attributed to SSL enabling the network to learn more discriminative features, such as shape, while learning the jigsaw task. Superior performance in the low data regime, and even the extra boost with 100\% labeled data, indicates that the jigsaw task effectively leverages unlabeled data. It is worth noting that the baseline outperforms SSL on the specificity metric. For our use case of precancerous lesion classification, sensitivity is more important than specificity, as missing precancerous lesions may lead to delayed treatment, a worse prognosis, and ultimately a reduced survival rate. 

\subsection{DOMAIN ADAPTATION}
\label{sec:res_da}
The goal of this experiment was to test the domain generalizability of semi-supervised learning. We train the model on labeled white light images (n=2326) and unlabeled NBI images(n=961), and then test the model using labeled NBI images (n=1685). The architecture, training protocol, and testing protocol remain the same as in the previous subsection \ref{sec:vld}. For the ablation study, the baseline model described in \ref{sec:vld} was used. For training the baseline model, we use the same set of white light labeled images (n=2326) as in SSL training. The hyperparameters used are an initial learning rate of 0.0001 and weight decay of 0.005 for both cases. In SSL, the number of jigsaw classes (P) was 100 and the unsupervised loss weight $\lambda = 1$ was used.

The results for the domain adaptation experiment are reported in \ref{tab:expt2}. To avoid any statistical error, we report the mean values for 3 runs initiated with different random seeds. We observe that the semi-supervised model exceeds the baseline in accuracy, F1 score and sensitivity by 1.92\%, 0.02 and 0.07 respectively. This superlative performance demonstrates that the semi-supervised methods take advantage of unlabeled target images to learn domain invariant feature representations. This may be enabled by the jigsaw puzzle solver learning the spatial correlation of images.

\begin{table}[]
\caption{\textbf{A comparison of Baseline and Jigsaw Pretext Semi-Supervised Learning for Domain Adaptation.}}
\label{tab:expt2}
\resizebox{\linewidth}{!}{%
\begin{tabular}{cccccc}
\hline\hline
                              & Accuracy       & F1 Score      & Sensitivity   & Specificity   & Precision     \\ \hline
\multicolumn{1}{l|}{Baseline} & 77.84\%          & 0.86          & 0.87          & \textbf{0.35} & \textbf{0.85} \\
\multicolumn{1}{l|}{SSL}      & \textbf{79.76\%} & \textbf{0.88} & \textbf{0.94} & 0.14          & 0.83         \\ \hline\hline
\end{tabular}%
}
\end{table}

\subsection{OUT-OF-DISTRIBUTION DETECTION}
In this experiment, we test semi-supervised performance as an out-of-distribution detector. In our problem setup, we treat white light images as in-distribution samples and NBI images as out-of-distribution. The SSL and baseline models and their training algorithms as lesion classifiers as described in \ref{sec:vld} was used in this experiment as well. The training set for the baseline and SSL consisted of 1921 labeled white light images, with the SSL model additionally used 1518 unlabeled WLI. We used the same hyperparameters as described in \ref{sec:res_da} for training the in-distribution models.

During inference, the out-of-distribution detector score $\kappa$ for the baseline is the KL-divergence between the prediction probabilities and uniform distribution. For SSL, we add the jigsaw loss to the KL-divergence term to compute $\kappa$ as described in equation \ref{equ:ood_score}. The test set consists of 416 white light images (label = 0) and 1685 NBI images (label = 1). The OOD scores and the labels are used to generate a Receiver Operator Characteristic (ROC) curve. The Area Under Receiver Operator Characteristic (AUROC) is then used as a metric to determine the efficacy of the OOD detector. The AUROC can be interpreted as the probability that the OOD score $\kappa$ for an out-of-distribution sample is greater than an in-distribution sample. 

\begin{figure}
  \centering
    \includegraphics[width=\linewidth]{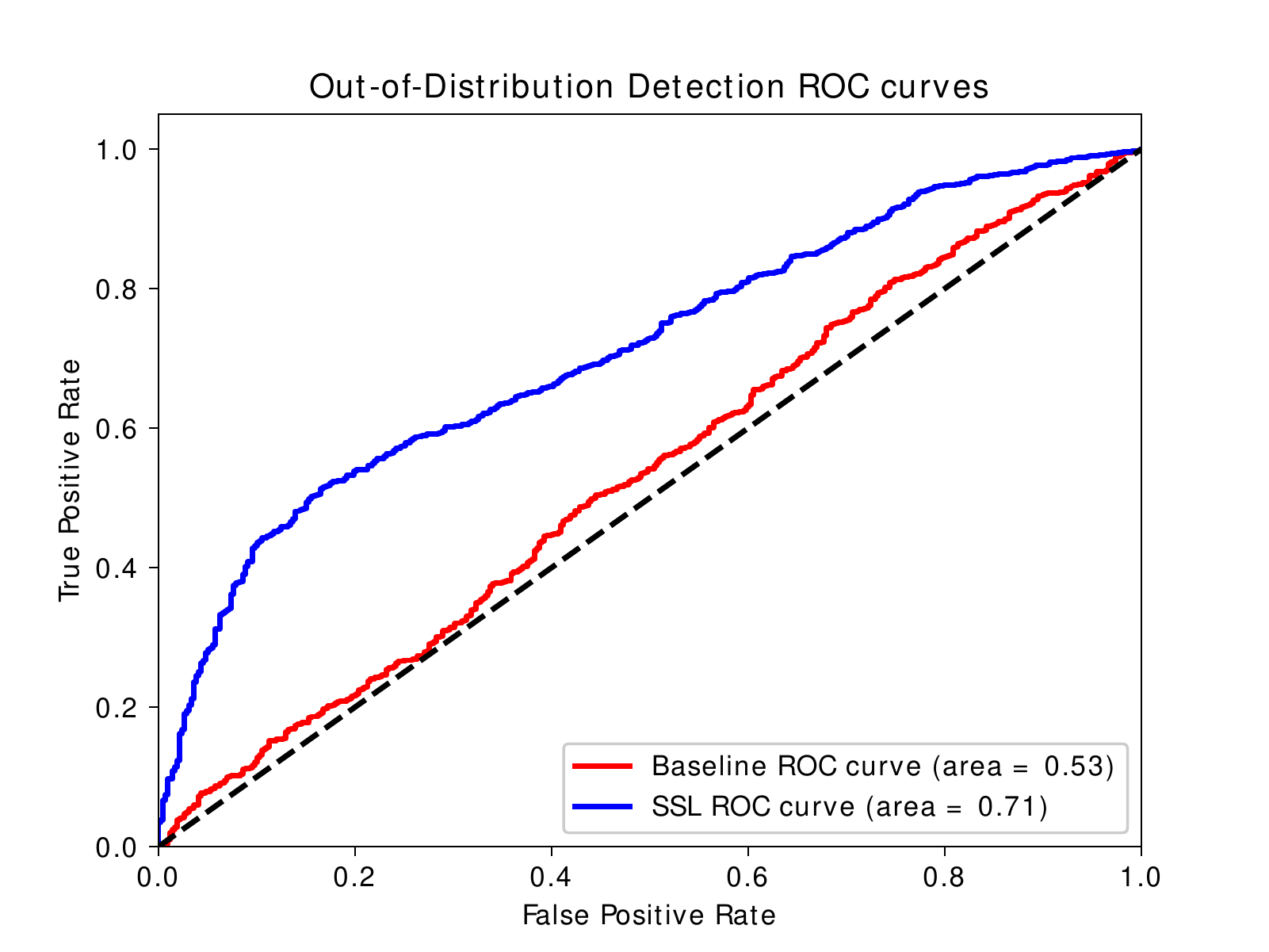}
  \caption{ROC curves for OOD detection comparing baseline and semi-supervised model}
  \label{fig:ood_roc}
\end{figure}

Figure \ref{fig:ood_roc} shows the results for OOD detection. A ROC curve for the model with median AUROC among three runs is reported. The SSL models has an AUROC of 0.71 as compared to 0.53 for the baseline. This shows that the unsupervised loss combines well with the KL-divergence term. The results demonstrate that attaching an auxiliary unsupervised head with a simplistic score can drastically improve the capability of the network as an OOD detector.

\section{CONCLUSION}

In this paper, we explore semi-supervised learning to utilize unlabeled data and improve lesion classification in colonoscopy images. We developed a phased training model using a jigsaw solving task and observed improved performance in metrics including accuracy and F1 score when compared with a purely supervised model. These data demonstrate that the addition of a jigsaw task helps the encoder generate discriminative features. We find that a semi-supervised learning model performs significantly better than a fully supervised method, especially in the low data regime. These results suggests that unsupervised learning is strongly regularizing the model. 

While the focus of semi-supervised learning works has traditionally been on accuracy metrics, in this paper we also study the effect of SSL on the generalizability and uncertainty of the model. In terms of generalizability, we show SSL's superior performance to supervised methods for domain adaptation. SSL improves performance on the target domain, using only unlabeled target distribution images. We also show that SSL models are better out-of-distribution detectors as compared to supervised models. This uncertainty measurement can simply be obtained from the prediction probabilities and jigsaw loss without requiring any architectural modifications. 

We would like to emphasize that the point of this study is \textit{not} to present the jigsaw based semi-supervised learning as the best-in-class model for the accuracy, domain adaptation, or OOD detection problems. Instead, we aim to establish proof-of-concept that adding an auxiliary semi-supervised task to supervised methods can significantly improve colonoscopy image analysis. In medical image analysis in general, the paucity of labeled data makes semi-supervised learning an important paradigm. Additionally, since domain generalization and out-of-distribution detection are important challenges in many practical clinical scenarios, semi-supervised learning holds significant promise to facilitate the translation of artificial intelligence techniques to real world applications. 

Future work to expand and further validate this general approach include exploring additional semi-supervised learning tasks such as image colorization, and patch prediction, or even a combination of these proxy tasks in a multi-task learning setup. To understand the dependence of the supervised objective on the semi-supervised learning proxy, the performance of a variety of colonoscopy challenges, such as polyp detection and segmentation, should be included, as well as additional proxy tasks. It is possible that the jigsaw task may not be optimal for improving the performance of lesion detection, for instance. The improvement in domain evaluation from SSL may be expanded by assessing not only across imaging modalities but also across different endoscopes with varying resolutions, illumination parameters, and frame rates. A deeper analysis of out-of-distribution detection, particularly for different types of out-of-distribution samples and the `harder' near distribution anomalies, is an important future step. Lastly, it would be valuable to explore how the SSL improvements change as the size of both the labeled and unlabeled datasets increase.

\bibliography{references} 

\begin{thebibliography}{10}

\bibitem{siegel2020}
R.~L. Siegel, K.~D. Miller, A.~Goding~Sauer, S.~A. Fedewa, L.~F. Butterly,
  J.~C. Anderson, A.~Cercek, R.~A. Smith, and A.~Jemal, ``Colorectal cancer
  statistics, 2020,'' {\em CA: A Cancer Journal for Clinicians}, vol.~70,
  no.~3, pp.~145--164, 2020.

\bibitem{joseph2016}
D.~A. Joseph, R.~G. Meester, A.~G. Zauber, D.~L. Manninen, L.~Winges, F.~B.
  Dong, B.~Peaker, and M.~van Ballegooijen, ``Colorectal cancer screening:
  estimated future colonoscopy need and current volume and capacity,'' {\em
  Cancer}, vol.~122, no.~16, pp.~2479--2486, 2016.

\bibitem{colucci2003}
P.~M. Colucci, S.~H. Yale, and C.~J. Rall, ``Colorectal polyps,'' {\em Clinical
  medicine \& research}, vol.~1, no.~3, pp.~261--262, 2003.

\bibitem{hayashi2013}
N.~Hayashi, S.~Tanaka, D.~G. Hewett, T.~R. Kaltenbach, Y.~Sano, T.~Ponchon,
  B.~P. Saunders, D.~K. Rex, and R.~M. Soetikno, ``Endoscopic prediction of
  deep submucosal invasive carcinoma: validation of the narrow-band imaging
  international colorectal endoscopic (nice) classification,'' {\em
  Gastrointestinal endoscopy}, vol.~78, no.~4, pp.~625--632, 2013.

\bibitem{li2014}
M.~Li, S.~M. Ali, S.~Umm-a OmarahGilani, J.~Liu, Y.-Q. Li, and X.-L. Zuo,
  ``Kudo’s pit pattern classification for colorectal neoplasms: a
  meta-analysis,'' {\em World Journal of Gastroenterology: WJG}, vol.~20,
  no.~35, p.~12649, 2014.

\bibitem{ijspeert2016}
J.~E. IJspeert, B.~A. Bastiaansen, M.~E. Van~Leerdam, G.~A. Meijer,
  S.~Van~Eeden, S.~Sanduleanu, E.~J. Schoon, T.~M. Bisseling, M.~C. Spaander,
  N.~Van~Lelyveld, {\em et~al.}, ``Development and validation of the wasp
  classification system for optical diagnosis of adenomas, hyperplastic polyps
  and sessile serrated adenomas/polyps,'' {\em Gut}, vol.~65, no.~6,
  pp.~963--970, 2016.

\bibitem{anderloni2014}
A.~Anderloni, M.~Jovani, C.~Hassan, and A.~Repici, ``Advances, problems, and
  complications of polypectomy,'' {\em Clinical and experimental
  gastroenterology}, vol.~7, p.~285, 2014.

\bibitem{he2019}
K.~He, H.~Fan, Y.~Wu, S.~Xie, and R.~Girshick, ``Momentum contrast for
  unsupervised visual representation learning,'' 2019.

\bibitem{chen2020}
T.~Chen, S.~Kornblith, M.~Norouzi, and G.~Hinton, ``A simple framework for
  contrastive learning of visual representations,'' 2020.

\bibitem{caron2020}
M.~Caron, I.~Misra, J.~Mairal, P.~Goyal, P.~Bojanowski, and A.~Joulin,
  ``Unsupervised learning of visual features by contrasting cluster
  assignments,'' 2020.

\bibitem{noroozi2016}
M.~Noroozi and P.~Favaro, ``Unsupervised learning of visual representations by
  solving jigsaw puzzles,'' in {\em European Conference on Computer Vision},
  pp.~69--84, Springer, 2016.

\bibitem{min2019overview}
J.~K. Min, M.~S. Kwak, and J.~M. Cha, ``Overview of deep learning in
  gastrointestinal endoscopy,'' {\em Gut and liver}, vol.~13, no.~4, p.~388,
  2019.

\bibitem{nogueira2020deep}
A.~Nogueira-Rodr{\'\i}guez, R.~Dom{\'\i}nguez-Carbajales,
  H.~L{\'o}pez-Fern{\'a}ndez, {\'A}.~Iglesias, J.~Cubiella, F.~Fdez-Riverola,
  M.~Reboiro-Jato, and D.~Glez-Pe{\~n}a, ``Deep neural networks approaches for
  detecting and classifying colorectal polyps,'' {\em Neurocomputing}, 2020.

\bibitem{mesejo2016computer}
P.~Mesejo, D.~Pizarro, A.~Abergel, O.~Rouquette, S.~Beorchia, L.~Poincloux, and
  A.~Bartoli, ``Computer-aided classification of gastrointestinal lesions in
  regular colonoscopy,'' {\em IEEE transactions on medical imaging}, vol.~35,
  no.~9, pp.~2051--2063, 2016.

\bibitem{zhang2016automatic}
R.~Zhang, Y.~Zheng, T.~W.~C. Mak, R.~Yu, S.~H. Wong, J.~Y. Lau, and C.~C. Poon,
  ``Automatic detection and classification of colorectal polyps by transferring
  low-level cnn features from nonmedical domain,'' {\em IEEE journal of
  biomedical and health informatics}, vol.~21, no.~1, pp.~41--47, 2016.

\bibitem{lui2019endoscopic}
T.~K. Lui, K.~K. Wong, L.~L. Mak, M.~K. Ko, S.~K. Tsao, and W.~K. Leung,
  ``Endoscopic prediction of deeply submucosal invasive carcinoma with use of
  artificial intelligence,'' {\em Endoscopy international open}, vol.~7, no.~4,
  p.~E514, 2019.

\bibitem{byrne2019real}
M.~F. Byrne, N.~Chapados, F.~Soudan, C.~Oertel, M.~L. P{\'e}rez, R.~Kelly,
  N.~Iqbal, F.~Chandelier, and D.~K. Rex, ``Real-time differentiation of
  adenomatous and hyperplastic diminutive colorectal polyps during analysis of
  unaltered videos of standard colonoscopy using a deep learning model,'' {\em
  Gut}, vol.~68, no.~1, pp.~94--100, 2019.

\bibitem{chen2018accurate}
P.-J. Chen, M.-C. Lin, M.-J. Lai, J.-C. Lin, H.~H.-S. Lu, and V.~S. Tseng,
  ``Accurate classification of diminutive colorectal polyps using
  computer-aided analysis,'' {\em Gastroenterology}, vol.~154, no.~3,
  pp.~568--575, 2018.

\bibitem{kandel2019su1741}
P.~Kandel, R.~LaLonde, V.~Ciofoaia, M.~B. Wallace, and U.~Bagci, ``Su1741
  colorectal polyp diagnosis with contemporary artificial intelligence,'' {\em
  Gastrointestinal Endoscopy}, vol.~89, no.~6, p.~AB403, 2019.

\bibitem{billah2017automatic}
M.~Billah, S.~Waheed, and M.~M. Rahman, ``An automatic gastrointestinal polyp
  detection system in video endoscopy using fusion of color wavelet and
  convolutional neural network features,'' {\em International journal of
  biomedical imaging}, vol.~2017, 2017.

\bibitem{mahmood2018multimodal}
F.~Mahmood, Z.~Yang, T.~Ashley, and N.~J. Durr, ``Multimodal densenet,'' {\em
  arXiv preprint arXiv:1811.07407}, 2018.

\bibitem{mahmood2019polyp}
F.~Mahmood, Z.~Yang, R.~Chen, D.~Borders, W.~Xu, and N.~J. Durr, ``Polyp
  segmentation and classification using predicted depth from monocular
  endoscopy,'' in {\em Medical Imaging 2019: Computer-Aided Diagnosis},
  vol.~10950, p.~1095011, International Society for Optics and Photonics, 2019.

\bibitem{ahmad2019artificial}
O.~F. Ahmad, A.~S. Soares, E.~Mazomenos, P.~Brandao, R.~Vega, E.~Seward,
  D.~Stoyanov, M.~Chand, and L.~B. Lovat, ``Artificial intelligence and
  computer-aided diagnosis in colonoscopy: current evidence and future
  directions,'' {\em The Lancet Gastroenterology \& Hepatology}, vol.~4, no.~1,
  pp.~71--80, 2019.

\bibitem{jing2019}
L.~Jing and Y.~Tian, ``Self-supervised visual feature learning with deep neural
  networks: A survey,'' 2019.

\bibitem{pathak2016context}
D.~Pathak, P.~Krahenbuhl, J.~Donahue, T.~Darrell, and A.~A. Efros, ``Context
  encoders: Feature learning by inpainting,'' in {\em Proceedings of the IEEE
  conference on computer vision and pattern recognition}, pp.~2536--2544, 2016.

\bibitem{vondrick2016generating}
C.~Vondrick, H.~Pirsiavash, and A.~Torralba, ``Generating videos with scene
  dynamics,'' in {\em Advances in neural information processing systems},
  pp.~613--621, 2016.

\bibitem{caron2018deep}
M.~Caron, P.~Bojanowski, A.~Joulin, and M.~Douze, ``Deep clustering for
  unsupervised learning of visual features,'' in {\em Proceedings of the
  European Conference on Computer Vision (ECCV)}, pp.~132--149, 2018.

\bibitem{doersch2015unsupervised}
C.~Doersch, A.~Gupta, and A.~A. Efros, ``Unsupervised visual representation
  learning by context prediction,'' in {\em Proceedings of the IEEE
  international conference on computer vision}, pp.~1422--1430, 2015.

\bibitem{oord2018representation}
A.~v.~d. Oord, Y.~Li, and O.~Vinyals, ``Representation learning with
  contrastive predictive coding,'' {\em arXiv preprint arXiv:1807.03748}, 2018.

\bibitem{gidaris2018unsupervised}
S.~Gidaris, P.~Singh, and N.~Komodakis, ``Unsupervised representation learning
  by predicting image rotations,'' {\em arXiv preprint arXiv:1803.07728}, 2018.

\bibitem{pathak2017learning}
D.~Pathak, R.~Girshick, P.~Doll{\'a}r, T.~Darrell, and B.~Hariharan, ``Learning
  features by watching objects move,'' in {\em Proceedings of the IEEE
  Conference on Computer Vision and Pattern Recognition}, pp.~2701--2710, 2017.

\bibitem{croitoru2017unsupervised}
I.~Croitoru, S.-V. Bogolin, and M.~Leordeanu, ``Unsupervised learning from
  video to detect foreground objects in single images,'' in {\em Proceedings of
  the IEEE International Conference on Computer Vision}, pp.~4335--4343, 2017.

\bibitem{ren2018cross}
Z.~Ren and Y.~Jae~Lee, ``Cross-domain self-supervised multi-task feature
  learning using synthetic imagery,'' in {\em Proceedings of the IEEE
  Conference on Computer Vision and Pattern Recognition}, pp.~762--771, 2018.

\bibitem{li2016unsupervised}
Y.~Li, M.~Paluri, J.~M. Rehg, and P.~Doll{\'a}r, ``Unsupervised learning of
  edges,'' in {\em Proceedings of the IEEE Conference on Computer Vision and
  Pattern Recognition}, pp.~1619--1627, 2016.

\bibitem{albuquerque2020improving}
I.~Albuquerque, N.~Naik, J.~Li, N.~Keskar, and R.~Socher, ``Improving
  out-of-distribution generalization via multi-task self-supervised
  pretraining,'' {\em arXiv preprint arXiv:2003.13525}, 2020.

\bibitem{korbar2018cooperative}
B.~Korbar, D.~Tran, and L.~Torresani, ``Cooperative learning of audio and video
  models from self-supervised synchronization,'' in {\em Advances in Neural
  Information Processing Systems}, pp.~7763--7774, 2018.

\bibitem{arandjelovic2017look}
R.~Arandjelovic and A.~Zisserman, ``Look, listen and learn,'' in {\em
  Proceedings of the IEEE International Conference on Computer Vision},
  pp.~609--617, 2017.

\bibitem{hendrycks2019}
D.~Hendrycks, M.~Mazeika, S.~Kadavath, and D.~Song, ``Using self-supervised
  learning can improve model robustness and uncertainty,'' 2019.

\bibitem{carlucci2019}
F.~M. Carlucci, A.~D'Innocente, S.~Bucci, B.~Caputo, and T.~Tommasi, ``Domain
  generalization by solving jigsaw puzzles,'' 2019.

\bibitem{hendrycks2016}
D.~Hendrycks and K.~Gimpel, ``A baseline for detecting misclassified and
  out-of-distribution examples in neural networks,'' 2016.

\bibitem{cheplygina2019not}
V.~Cheplygina, M.~de~Bruijne, and J.~P. Pluim, ``Not-so-supervised: a survey of
  semi-supervised, multi-instance, and transfer learning in medical image
  analysis,'' {\em Medical image analysis}, vol.~54, pp.~280--296, 2019.

\bibitem{iglesias2010agreement}
J.~E. Iglesias, C.-Y. Liu, P.~Thompson, and Z.~Tu, ``Agreement-based
  semi-supervised learning for skull stripping,'' in {\em International
  Conference on Medical Image Computing and Computer-Assisted Intervention},
  pp.~147--154, Springer, 2010.

\bibitem{meier2014patient}
R.~Meier, S.~Bauer, J.~Slotboom, R.~Wiest, and M.~Reyes, ``Patient-specific
  semi-supervised learning for postoperative brain tumor segmentation,'' in
  {\em International Conference on Medical Image Computing and
  Computer-Assisted Intervention}, pp.~714--721, Springer, 2014.

\bibitem{dittrich2014spatio}
E.~Dittrich, T.~R. Raviv, G.~Kasprian, R.~Donner, P.~C. Brugger, D.~Prayer, and
  G.~Langs, ``A spatio-temporal latent atlas for semi-supervised learning of
  fetal brain segmentations and morphological age estimation,'' {\em Medical
  image analysis}, vol.~18, no.~1, pp.~9--21, 2014.

\bibitem{gu2017semi}
L.~Gu, Y.~Zheng, R.~Bise, I.~Sato, N.~Imanishi, and S.~Aiso, ``Semi-supervised
  learning for biomedical image segmentation via forest oriented super pixels
  (voxels),'' in {\em International Conference on Medical Image Computing and
  Computer-Assisted Intervention}, pp.~702--710, Springer, 2017.

\bibitem{bai2017semi}
W.~Bai, O.~Oktay, M.~Sinclair, H.~Suzuki, M.~Rajchl, G.~Tarroni, B.~Glocker,
  A.~King, P.~M. Matthews, and D.~Rueckert, ``Semi-supervised learning for
  network-based cardiac mr image segmentation,'' in {\em International
  Conference on Medical Image Computing and Computer-Assisted Intervention},
  pp.~253--260, Springer, 2017.

\bibitem{wang_focalmix:_2020}
D.~Wang, Y.~Zhang, K.~Zhang, and L.~Wang, ``{FocalMix}: {Semi}-{Supervised}
  {Learning} for {3D} {Medical} {Image} {Detection},'' in {\em 2020
  {IEEE}/{CVF} {Conference} on {Computer} {Vision} and {Pattern} {Recognition}
  ({CVPR})}, (Seattle, WA, USA), pp.~3950--3959, IEEE, June 2020.

\bibitem{baur2019fusing}
C.~Baur, B.~Wiestler, S.~Albarqouni, and N.~Navab, ``Fusing unsupervised and
  supervised deep learning for white matter lesion segmentation,'' in {\em
  International Conference on Medical Imaging with Deep Learning}, pp.~63--72,
  2019.

\bibitem{diaz-pinto_retinal_2019}
A.~Diaz-Pinto, A.~Colomer, V.~Naranjo, S.~Morales, Y.~Xu, and A.~F. Frangi,
  ``Retinal {Image} {Synthesis} and {Semi}-{Supervised} {Learning} for
  {Glaucoma} {Assessment},'' {\em IEEE Transactions on Medical Imaging},
  vol.~38, pp.~2211--2218, Sept. 2019.

\bibitem{ahsan_video_2019}
U.~Ahsan, R.~Madhok, and I.~Essa, ``Video {Jigsaw}: {Unsupervised} {Learning}
  of {Spatiotemporal} {Context} for {Video} {Action} {Recognition},'' in {\em
  2019 {IEEE} {Winter} {Conference} on {Applications} of {Computer} {Vision}
  ({WACV})}, (Waikoloa Village, HI, USA), pp.~179--189, IEEE, Jan. 2019.

\bibitem{zhuang2019self}
X.~Zhuang, Y.~Li, Y.~Hu, K.~Ma, Y.~Yang, and Y.~Zheng, ``Self-supervised
  feature learning for 3d medical images by playing a rubik’s cube,'' in {\em
  International Conference on Medical Image Computing and Computer-Assisted
  Intervention}, pp.~420--428, Springer, 2019.

\bibitem{noroozi_boosting_2018}
M.~Noroozi, A.~Vinjimoor, P.~Favaro, and H.~Pirsiavash, ``Boosting
  {Self}-{Supervised} {Learning} via {Knowledge} {Transfer},'' in {\em 2018
  {IEEE}/{CVF} {Conference} on {Computer} {Vision} and {Pattern}
  {Recognition}}, (Salt Lake City, UT), pp.~9359--9367, IEEE, June 2018.

\bibitem{taleb2019multimodal}
A.~Taleb, C.~Lippert, T.~Klein, and M.~Nabi, ``Multimodal self-supervised
  learning for medical image analysis,'' {\em arXiv preprint arXiv:1912.05396},
  2019.

\bibitem{tao2020revisiting}
X.~Tao, Y.~Li, W.~Zhou, K.~Ma, and Y.~Zheng, ``Revisiting rubik's cube:
  Self-supervised learning with volume-wise transformation for 3d medical image
  segmentation,'' {\em arXiv preprint arXiv:2007.08826}, 2020.

\bibitem{ross2018exploiting}
T.~Ross, D.~Zimmerer, A.~Vemuri, F.~Isensee, M.~Wiesenfarth, S.~Bodenstedt,
  F.~Both, P.~Kessler, M.~Wagner, B.~M{\"u}ller, {\em et~al.}, ``Exploiting the
  potential of unlabeled endoscopic video data with self-supervised learning,''
  {\em International journal of computer assisted radiology and surgery},
  vol.~13, no.~6, pp.~925--933, 2018.

\bibitem{hendrycks2016baseline}
D.~Hendrycks and K.~Gimpel, ``A baseline for detecting misclassified and
  out-of-distribution examples in neural networks,'' {\em arXiv preprint
  arXiv:1610.02136}, 2016.

\bibitem{hendrycks2018deep}
D.~Hendrycks, M.~Mazeika, and T.~Dietterich, ``Deep anomaly detection with
  outlier exposure,'' {\em arXiv preprint arXiv:1812.04606}, 2018.

\bibitem{NEURIPS2019_9015}
A.~Paszke, S.~Gross, F.~Massa, A.~Lerer, J.~Bradbury, G.~Chanan, T.~Killeen,
  Z.~Lin, N.~Gimelshein, L.~Antiga, A.~Desmaison, A.~Kopf, E.~Yang, Z.~DeVito,
  M.~Raison, A.~Tejani, S.~Chilamkurthy, B.~Steiner, L.~Fang, J.~Bai, and
  S.~Chintala, ``Pytorch: An imperative style, high-performance deep learning
  library,'' in {\em Advances in Neural Information Processing Systems 32}
  (H.~Wallach, H.~Larochelle, A.~Beygelzimer, F.~d\textquotesingle
  Alch\'{e}-Buc, E.~Fox, and R.~Garnett, eds.), pp.~8024--8035, Curran
  Associates, Inc., 2019.

\bibitem{he2016}
K.~He, X.~Zhang, S.~Ren, and J.~Sun, ``Deep residual learning for image
  recognition,'' in {\em Proceedings of the IEEE conference on computer vision
  and pattern recognition}, pp.~770--778, 2016.

\bibitem{kingma2014method}
D.~P. Kingma and J.~Ba, ``Adam: A method for stochastic optimization,'' 2014.
\newblock 3rd International Conference for Learning Representations, San Diego,
  2015.

\end{thebibliography}
\bibliographystyle{ieeetr}

\EOD

\end{document}